\documentclass[wcp]{jmlr}

\usepackage{longtable}% for long tables

\usepackage{amsbsy}
\usepackage{amsmath}
\usepackage{amssymb}
\usepackage{graphicx}

\usepackage{setspace}
\usepackage{fancyhdr}

\usepackage{relsize} %big sum

\usepackage{sidecap} %for SC figure

\usepackage{multicol} %for multicolumn
\usepackage{multirow}
\usepackage{hyperref}
\usepackage{color}

\newcommand{\argmax}{\mathop{\rm argmax}\limits}

\usepackage{stmaryrd}

\usepackage{booktabs}
\usepackage{amsmath,amssymb,amsfonts}
\usepackage{algorithm,algorithmic}
\usepackage{graphicx}  %Required
\usepackage{multirow}
\usepackage[export]{adjustbox}% http://ctan.org/pkg/adjustbox

\jmlrworkshop{\textsc{SIAM Data Mining 2019}}

\begin{document}
\pagestyle{plain}

\pagestyle{fancy}
\fancyhead{}
\fancyhead[C]{\textsc{\small Monte-Carlo Single Rooted DAG Search by Best Arm Identification}}
\renewcommand{\headrulewidth}{0pt}

\title[Monte-Carlo Single Rooted DAG Search by Best Arm Identification]{Feature selection as Monte-Carlo Search in Growing Single Rooted Directed Acyclic Graph by Best Leaf Identification}

 \author{\Name{Aur\'elien P\'elissier} \Email{aurelien.pelissier@ens-paris-saclay.fr}\\
  \addr \'Ecole Normale Sup\'erieure Paris-Saclay, France\\
  \Name{Atsuyoshi Nakamura} \Email{atsu@ist.hokudai.ac.jp }\\
  \Name{Koji Tabata} \Email{ktabata@es.hokudai.ac.jp  }\\
  \addr Graduate school of Information Science and Technology, Hokkaido University, Japan
 }

\maketitle

\fancyfoot[R]{\scriptsize{Copyright \textcopyright\ 2019 by SIAM\\
Unauthorized reproduction of this article is prohibited}}

\begin{abstract} \small\baselineskip=9pt Monte Carlo tree search (MCTS) has received considerable interest due to its spectacular success in the difficult problem of computer Go and also proved beneficial in a range of other domains. A major issue that has received little attention in the MCTS literature is the fact that, in most games, different actions can lead to the same state, that may lead to a high degree of redundancy in tree representation and unnecessary additional computational cost. We extend MCTS to single rooted directed acyclic graph (SR-DAG), and consider the Best Arm Identification (BAI) and the Best Leaf Identification (BLI) problem of an expanding SR-DAG of arbitrary depth. We propose algorithms that are ($\varepsilon$, $\delta$)-correct in the fixed confidence setting, and prove an asymptotic upper bounds of sample complexity for our BAI algorithm. As a major application for our BLI algorithm, a novel approach for Feature Selection is proposed by representing the feature set space as a SR-DAG and repeatedly evaluating feature subsets until a candidate for the best leaf is returned, a proof of concept is shown on benchmark data sets.\\

\vspace{-0.2cm}

\hspace{-18pt}\textbf{Keywords:} DAG, MCTS, BAI, BLI, Feature Selection\end{abstract}

\section{Introduction}The combination of Monte Carlo tree search (MCTS) with bandit strategies has proven remarkably efficient and has received considerable interest due to its spectacular success, mainly in the difficult problem of computer Go, but also in a wide range of other domains (eg. Optimization, Scheduling, ..) [\cite{browne2012survey}].

An  important  issue  that  has  not  seen  much  attention  in  the  MCTS  literature  is  the fact that, in most games, different actions can lead to the same state (usually referred as transpositions), or states can be revisited [\cite{gusmao2012towards}]. Thus, the game's state space should be represented as a connected graph, not as a tree. In the \textit{basic} well-known UCT algorithm [\cite{kocsis2006bandit}], the space of the game is treated as a tree and this can lead to having multiple nodes for the same position, resulting in a high degree of redundancy and thus requiring more episodes to get accurate estimates of action values.

By representing the game space as a connected graph rather than a tree, and defining an appropriate bandit policy, we hope to share the knowledge between different paths that lead to the same state and gain efficiency in the Monte-Carlo search. A straight forward application would reside in strategy game computational problems such as Go or Hex, where the consideration for transpositions has already proven to significantly improve the performances of game agents [\cite{saffidine2012ucd}]. But other concrete examples, such as Feature Selection problems that can be formalized as a reinforcement learning problem [\cite{gaudel2010feature}], could gain considerably from such improvements. Because the corresponding game state space of feature selection problems contains a very high number of transpositions.

Another important key concept in Monte Carlo Search is its expansion policy, that expands the tree inhomogeneously depending on the empirical means of the nodes. The well-known UCT algorithm holds its success from its expansion policy, because it allows the Monte Carlo Search to get more accurate predictions on the most promising arms by exploring deeper in the tree.

In this work we consider the game space as a single rooted directed acyclic graph (SR-DAG) and focus on the best arm identification problem at the root by Monte Carlo search. We start by describing the Monte Carlo Search by Best Arm Identification in Expanding DAG (BAI-exMCDS) generic architecture, that relies on confidence intervals on the node values in order to solve the BAI problem in a DAG of arbitrary depth. The proposed architecture also deals with growing DAGs by introducing an expansion policy that adds nodes depending of the number of leaf evaluations. We propose the LUCB-exMCDS algorithm that returns an ($\varepsilon$, $\delta$)-correct solution to the BAI problem with sample complexity guaranty.

The second part deals with the Best Leaf Identification (BLI) problem by solving the BAI problem at multiple stages in the SR-DAG, with the returned recommended leaf being ($\varepsilon$, $\delta$)-correct. Finally, as a major application for our algorithm, we consider Feature Selection as a BLI problem and a proof of concept is shown on benchmark data sets.\\

\hspace{-18pt}\textbf{Related work }$\ $ Bandits models, where an agent repeatedly selects one out of several arms and receives a reward generated according to some distribution depending on the selected arm, have been studied since the 1930s [\cite{thompson1933likelihood}]. They are usually solved in the \textit{regret minimization setting} (UCB algorithm [\cite{auer2002finite}]), but other variants, like the best arm identification (BAI) problem are also considered. The BAI problem consists of quickly and accurately finding the arm with highest mean. Two major algorithms are lil'UCB [\cite{jamieson2014lil}] and UGapE [\cite{gabillon2012best}] that solve the BAI problem in both \textit{fixed confidence} and \textit{fixed budget} setting.

In the context of MCTS, the BAI problem corresponds to identifying the next best action to take at the root. While the popular UCT algorithm [\cite{kocsis2006bandit}] has successfully adapted bandit strategies to growing tree search, there are only very weak theoretical guarantees for UCT. \cite{kaufmann2017monte} and \cite{huang2017structured} recently introduced novel algorithms for the BAI problem in MCTS with fixed trees that are based on LUCB [\cite{kalyanakrishnan2012pac}] and UGapE. Their algorithms are similar and differs only by the way the most promising arm \textit{bt} is picked at the root. Both have shown good sample complexity guaranty in the \textit{fixed confidence} setting. However, their work is currently limited to fixed tree, which is considerably limiting the efficiency of game search.

Regarding the BLI problem, UCT can recommend a candidate for the best leaf by returning the path with maximal average reward or the most often visited path in the search tree. There has however not been much attention to the BLI problem in the MCTS literature, mainly because most applications of MCTS only concern the next move to take from the current node. Feature selection can be formalized as a BLI problem and \cite{gaudel2010feature} proposed the algorithm  FUSE, that  starts  from  the empty feature set and relies on UCT to identify the best feature subset.

Finally, since most-space game are better represented by a DAG rather than a tree, there have been attempts to generalize MCTS to connected graphs. \cite{saffidine2012ucd} show that the task is difficult and that unwanted behavior are likely to occur if the backpropagation or bandit policies are not carefully considered (the algorithm does not converge to optimal action), they attempted to extend UCT by introducing the upper confidence bound for rooted DAGs (UCD) algorithm ; modification to UCT using transposition tables was also proposed by \cite{childs2008transpositions}. Both of these proposals lack of theoretical guaranty and to our current knowledge, there is no existing algorithm for Monte Carlo search in growing DAG with theoretical guaranty. While the algorithm of \cite{huang2017structured} allows transpositions, it does not deal with the non-uniqueness of the optimal arm, that is a common fact in DAGs.\\

\hspace{-18pt}\textbf{Our contribution }$\ $  In this paper, we extend the BAI-MCTS architecture proposed by \cite{kaufmann2017monte} to expanding SR-DAG by introducing an expansion rule that adds nodes depending on the number of leaf evaluations, as well as an addition rule that selects the node to be added accounting for the information gained during the search. We prove that our algorithm is $(\varepsilon, \delta)$-correct and give an asymptotic upper bound of its sample complexity. We also give a solution to the BLI problem with $(\varepsilon, \delta)$-correctness and introduce a novel algorithm for the Feature Selection problem with theoretical guaranties by representing the feature set space as a SR-DAG.

\section{Best Arm Identification in Growing Monte Carlo SR-DAG Search}

\subsection{SR-DAG formalism and BAI-exMCDS}

This section retakes the previous work presented by \cite{kaufmann2017monte} on BAI-MCTS and extends it to growing SR-DAGs. Even though there is high similarities between MCDS and MCTS, some differences that fundamentally change the behaviors of the search have to be considered. To ease the presentation, we focus on single player game, but our algorithm can be easily extended to two player interactions by representing the sequence of possible successive moves with a Min-Max game graph as in \cite{teraoka2014efficient}.

\subsubsection{Monte-Carlo Search in expanding DAGs (exMCDS)}
We consider a fixed connected \emph{DAG} (Directed Acyclic Graph) $\mathcal{D}=(\mathcal{V},\mathcal{E})$ composed of \emph{nodes} $s \in \mathcal{V}$ related to each other by \emph{directed edges} $(s_1,s_2)\in\mathcal{E}\subseteq \mathcal{V}\times\mathcal{V}$.
For an edge $(s_1,s_2)\in \mathcal{E}$, $s_1$ is said to be a \emph{parent} of $s_2$ and $s_2$ is said to be a \emph{child} of $s_1$. Node $s_1$ is said to be an \emph{ancestor} of node $s_2$ if there is a directed path from $s_1$ to $s_2$. For each node $s \in \mathcal{V}$, we denote by $\mathcal{C}(s)$ the set of its children and by $\mathcal{P}(s)$ the set of its parents. The \emph{root} $s_0 \in \{ s \in \mathcal{D} \ | \ \mathcal{P}(s) = \varnothing \}$ is assumed unique, and we finally introduce the \emph{terminal leaf} node set $\mathcal{L} = \{ s \in \mathcal{D} \ | \ \mathcal{C}(s) = \varnothing \}$. Note that the major difference of a DAG compared to a tree is that the parent set $\mathcal{P}(s)$ may contains more than one node.

Let $\mathcal{D}_t=(\mathcal{V}_t,\mathcal{E}_t)$ be the SR-DAG at time $t$ ,
which is a connected subgraph of $\mathcal{D}$ containing the root $s_0$, that is, $\mathcal{V}_t\subseteq \mathcal{V}$,
$\mathcal{E}_t\subseteq \mathcal{E}$, $s_0\in \mathcal{V}_0$, and for each $s\in \mathcal{V}_t$ there is a path $s_0s_1\dots s_n$ such that $(s_i,s_{i+1})\in \mathcal{E}_t$ ($i=0,\dots,n-1$) and $s_n=s$.  Let $\mathcal{L}_0$ be the initial leaf node set and $\mathcal{L}_t$ the temporary leaf node set at step $t$.

For each terminal leaf node $\ell$, we assume a stochastic oracle $\mathcal{O}_\ell$ that returns a value $X\in [0,1]$ generated according to an unknown distribution over $[0,1]$ with mean $\mu_{\ell}$ for each call. We also introduce an intermediate stochastic oracle $^{*}\mathcal{O}_{\ell}$ that evaluates the temporary leaves $\ell_t \in  \mathcal{L}_t \setminus (\mathcal{L}_t \cap \mathcal{L})$,  which also returns a value $X\in [0,1]$ generated according to an unknown distribution over $[0,1]$ with mean $^*\mu_{\ell_t}$ for each call. One should note that oracles $\mathcal{O}_\ell$ and $^{*}\mathcal{O}_{\ell_t}$ can be very different. As an example, in the problem of computer Go, terminal leaves are representing wins and losses, and thus only takes values 0 and 1 ($\mu_\ell \in \{ 0,1 \}$). On the other hand, intermediate leaves values correspond to the probability of winning from the current game state when playing random moves until the end ($^*\mu_{\ell_t} \in [0,1]$).\\

The value $V(s)$ for any node $s \in \mathcal{V}$ is recursively defined with
\begin{equation}
\nonumber
    V(s) = \mu_s \ \text{ if } s \in \mathcal{L}, \hspace{0.6cm} V(s) = \max_{c \in \mathcal{C}(s)} \hspace{0.1cm} V(c) \ \text{ otherwise.}
\end{equation}
The best child of the root $s^*$ is the root's child with highest value,
\begin{equation}
  \nonumber
    s^{*} = \argmax_{s \in \mathcal{C}(s_0)}V(s).
\end{equation}

The \emph{MCDS algorithm} is an algorithm that sequentially selects paths from the root to a leaf in the DAG and calls the corresponding leaf oracle  $\mathcal{O}_\ell$ or $^*\mathcal{O}_\ell$ to collect a sample of the leaf to identify $s^*$. 

\subsubsection{$(\varepsilon, \delta)$-PAC learning framework}

In the \emph{PAC} (Probably Approximately Correct) learning frameworks,
an algorithm is said to be $(\varepsilon, \delta)$-correct if,
for given a risk level $0 < \delta < 1$ and some accuracy parameter $\varepsilon \geq 0$,
it outputs $\hat{s_\tau} \in \mathcal{C}(s_0)$ whose value is within $\varepsilon$ of the value of the best root's child $s^*$, with probability at least $1 - \delta$:
$$\mathbb{P} \left[ V(s^*) - V(\hat{s_\tau}) \leq \varepsilon \right] \geq 1 - \delta.$$
Our study partially aims at designing an algorithm that uses as few leaf evaluation $\tau$ as possible to fulfill this condition.

\subsubsection{Confidence interval and representative nodes}
\label{confidenceInterval}
For each leaf $\ell \in \mathcal{L}_t$, a \emph{confidence interval} $\mathcal{I}_\ell(t)$ at time $t$ is built using the past observations from this leaf, with $\text{U}_\ell(t)$ (resp. $\text{L}_\ell(t)$) being an \emph{upper confidence bound} (resp. a \emph{lower confidence bound}) of the value $V(\ell) = \mu_\ell$ at time $t$:
$$\mathcal{I}_\ell(t) = [\text{L}_\ell(t), \text{U}_\ell(t)].$$
These confidence intervals are then propagated upwards in the SR-DAG. For each internal node $s \in \mathcal{D}_t \setminus \mathcal{L}_t$ , the interval confidence $\mathcal{I}_s(t) = [\text{L}_s(t),\text{U}_s(t)]$ is recursively defined with
$$\text{L}_s(t) = \max_{c \in \mathcal{C}(s)} \text{L}_c(t), \hspace{1cm} \text{U}_s(t) = \max_{c \in \mathcal{C}(s)} \text{U}_c(t).$$
A \emph{representative child} $c_s(t)$ of node $s$ is introduced as
$$c_s(t)  = \argmax_{c \in \mathcal{C}(s)} \text{U}_c(t),$$
and a \textit{representative leaf} $\ell_s(t)$ of node $s$ is defined recursively as follows:
\begin{equation}
\nonumber
    \ell_s(t) = s \text{ if } s \in \mathcal{L}_t, \hspace{1cm} \ell_s(t) = \ell_{c_s(t)}(t) \text{ otherwise.}
\end{equation}

\begin{figure}[h!t]
    \centering
    \includegraphics[width=0.45\columnwidth]{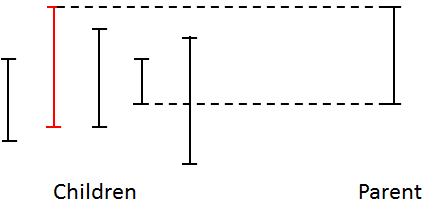}
    \caption{Construction of confidence interval and representative child (in red). }
    \label{fig:rep}
\end{figure}

The construction of the confidence intervals and associated representative children are illustrated in Figure\ref{fig:rep}. Since the confidence intervals of each nodes represent their plausible values, the representative child can be interpreted as an \textit{optimistic} move.

\subsubsection{Confidence interval choice}
\label{exploration_beta}
Denoting $N_\ell(t)$ the number of evaluations of leaf $\ell$ after $t$ steps, \cite{kaufmann2017monte} and \cite{huang2017structured} introduced the following confidence intervals to have a $(\varepsilon,\delta)$-correct algorithm, that holds for $\delta \leq \min (1, 0.1|\mathcal{L}|)$:
    \begin{equation}
    \label{exploration_eq}
        \begin{split}
            &\text{L}_\ell(t) = \hat{\mu_\ell}(t) - \sqrt{\frac{\beta(N_\ell(t), \delta)}{2 N_\ell (t)}} \ \ \ \text{and} \ \ \ \text{U}_\ell(t) = \hat{\mu_\ell}(t) + \sqrt{\frac{\beta(N_\ell(t), \delta)}{2 N_\ell (t)}}\\
            &\text{with} \ \ \ \beta(N,\delta) = \ln \left(\frac{|\mathcal{L}|}{\delta}\right) + 3 \ln \ln\left(\frac{|\mathcal{L}|}{\delta}\right) + \frac{3}{2} \ln \left(\ln(N) + 1 \right)
        \end{split}
    \end{equation}
    where $\hat{\mu_\ell}(t)$ is the sample mean of $N_\ell(t)$ samples obtained from oracle $\mathcal{O}_\ell$.
    
The union bound over $|\mathcal{L}_t|$ makes the exploration function over-conservative and in practice \cite{kaufmann2017monte} recommend the use of $\beta(\delta,N) = \ln \left(\frac{\ln (eN)}{\delta}\right)$. The differences between these two exploration functions are discussed in the experimental section of \textit{supplementary materials} (Section \ref{experiment}).

\subsubsection{BAI-exMCDS architecture $\ $}
The BAI-exMCDS architecture class combines a BAI algorithm from the root with an exploration of the DAG based on confidence intervals on the node values as well as an expansion policy that add new leaves to the DAG to get more accurate estimations of the root's child values, it combines five ingredients:

\begin{itemize}
    \item The \textbf{sampling rule} $\texttt{BAISelect} (s_0,\mathcal{D}_t)$ that select a child from node $s_0$.
    \item The \textbf{stopping rule} $\texttt{BAIStop} (s_0,\mathcal{D}_t)$ that return \texttt{True} if the algorithm decides to stop.
    \item The \textbf{recommendation rule} $\texttt{BAIReco} (s_0,\mathcal{D}_t)$ that select a candidate for the best child of $s_0$.
    \item The \textbf{expansion rule} $\texttt{BAIExpand}(\mathcal{D}_t,t)$ that returns \texttt{True} if the algorithm decides to expand the DAG at step $t$.
    \item The \textbf{addition rule} $\texttt{BAIAdd}(\mathcal{D}_t,\mathcal{D})$ that selects a new node $s \in \mathcal{V}\setminus \mathcal{V}_t$ to be added to the DAG such as $\exists s_p \in \mathcal{V}_t$ for which $s \in \mathcal{C}(s_p)$, together will all edges $(s_p',s)$,  $\forall s_p'\in \mathcal{P}(s) \cap \mathcal{V}_t$.
\end{itemize}

Since the BAI problem in MCDS refers to the root's children, the sampling rule uses the information about depth-one nodes $(\mathcal{C}(s_0))$, that has to be updated at the end of each round.
\begin{algorithm2e}
\renewcommand{\thealgocf}{BAI-exMCDS}
\caption{Basic architecture of BAI-exMCDS}\label{MCDS}
\SetKwInOut{Input}{input}
\SetKwInOut{Output}{output}
    \Input{sr-dags $\mathcal{D}_0$, $\mathcal{D}$, risk level $\delta$, accuracy $\varepsilon$, expansion parameter $b$}
    \Output{recommended arm $\hat{s}^*$}
    $\mathcal{D}_t\gets \mathcal{D}_0$, \ $ t \gets  0$\\
    \While{\textbf{not} \emph{\texttt{BAIStop}$(s_0,\mathcal{D}_t)$}}{
        \eIf{ \emph{\texttt{BAIExpand}}$(\mathcal{D}_t,t)$}{
            $\mathcal{D}_{t+1}\gets \texttt{BAIAdd}(\mathcal{D}_t,\mathcal{D})$
        }{
            $\mathcal{D}_{t+1}\gets \mathcal{D}_t$
        }
        $R_{t+1} \gets \texttt{BAISelect}(s_0,\mathcal{D}_{t+1})$\\
        
        \eIf{ $\ell_{R_{t+1}} \in \mathcal{L}$}{
            $X\gets \mathcal{O}_{\ell_{R_{t+1}}}$
        }{
            $X\gets ^{*}\mathcal{O}_{\ell_{R_{t+1}}}$
        }
        Update the confidence interval of $\ell_{R_{t+1}(t)}$\\
        Update the confidence intervals for all the ancestors of $\ell_{R_{t+1}}(t)$.\\
        $t \gets t + 1$
}
    \textbf{return} $\texttt{BAIReco}(s_0,\mathcal{D}_t)$
%\end{algorithmic}
\end{algorithm2e}

\hspace{-18pt}\textit{\textbf{Remark }A major change arising from considering a DAG over a tree is that the number of nodes to be updated after leaf evaluation scales exponentially with the depth of that leaf (because the number of parent per node is in average higher than 1).}\\

This \emph{exponential backpropagation} is in contrast with the \emph{linear backpropagation} that is traditionally used in MCTS. With exponential backpropagation, we hope to update the information in a more productive way that may considerably improve the efficiency of the search. 

\cite{saffidine2012ucd} study this difference in the backpropagation method as the \textit{updated-all} vs \textit{updated-descent} rule, and they demonstrate the risk of using the \textit{updated-all} policy with UCT (algorithm that does not converge to optimal root's child). However, while UCT updates confidence intervals for internal nodes by propagating a reward, BAI-MCDS updates them by propagating confidence bounds of leaves, and does not propagate the reward itself, thus such problems cannot occur.\\

\subsection{LUCB-exMCDS}
In reference to the LUCB-MCTS ($\varepsilon$, $\delta$)-correct algorithm described by~\cite{kaufmann2017monte}, we introduce LUCB-exMCDS that extends BAI search to expanding single rooted DAGs.

\subsubsection{Best Arm identification policy}

The main difference arising from considering a DAG rather than a tree is that different children from the same parent node may have the same \textit{representative leaf} in DAGs, whereas it is never the case in trees, and as the result the uniqueness of the best arm assumption may not hold. The stopping rule and sampling rule are modified accordingly to account for this possibility.\\

For each root's child $a \in \mathcal{C}(s_0)$, we introduce the set $\overline{\mathcal{C}}_a(t)$, which contains the root's children whose \textit{representative leaf} at time $t$ is different from that of $a$,
\begin{equation}
\nonumber
    \overline{\mathcal{C}}_a(t) = \{ s' \in \mathcal{C}(s_0) \ | \ \ell_{s'}(t) \neq \ell_a(t) \}.
\end{equation}
    
If all the children from the root node have the same representative leaf, then any child can be considered as the best arm and the BAI problem becomes trivial. In the following we assume that $\forall s \in \mathcal{C}(s_0), \ \overline{\mathcal{C}}_s(t) \neq \emptyset$. 

\paragraph{\texttt{BAISelect}$(s_0,\mathcal{D}_t)$:} Return the node's child $R_{t+1}$ selected as follows:
\begin{align*}
    &b_t \gets \argmax_{s \in \mathcal{C}(s_0)}\ \text{L}_s(t), \hspace{20pt} c_t \gets \argmax_{s \in \overline{\mathcal{C}}_{b_t}(s_0)} \text{U}_s(t),\\
    &R_{t+1} \gets \argmax_{s \in \{b_t,c_t\}} [\text{U}_s(t) - \text{L}_s(t)]. 
\end{align*}    
\paragraph{\texttt{BAIStop}$(s_0,\mathcal{D}_t)$:} Return \texttt{True} if and only if $( \text{U}_{c_t(t)} - \text{L}_{b_t}(t) < \varepsilon) $.\\
\paragraph{\texttt{BAIReco}$(s_0,\mathcal{D}_t)$:} return $b_t$.\\

$b_t$ represents a guess for the best depth-one node, while $c_t$ is an optimistic challenger that has the maximal possible value among the other depth-one nodes. Both nodes need to be explored enough in order to find the best arm quickly.\\

\hspace{-18pt}\textit{\textbf{Remark }This algorithm is closer to the UGapE\hspace{1pt}-\hspace{1pt}MCTS algorithm than the LUCB\hspace{1pt}-\hspace{1pt}MCTS algorithm described by \cite{kaufmann2017monte} because it relies only on confidence intervals rather than empirical means of the leaves. In the original UGapE\hspace{1pt}-\hspace{1pt}MCTS algorithm, $b_t$ is defined as the node maximizing its index $B_s = \max_{s' \neq s} [\text{U}_{s'}(t) - \text{L}_s(t)]$. However, that definition seemed unnecessary complicated for our problem, and inspired from \cite{huang2017structured}, we rather chose a simpler definition for our arm selection policy at the root (the one maximizing its lower-bound).}

\subsubsection{Expansion policy }

\label{expansion_rule}

Intuitively, because the sample complexity of LUCB-exMCDS scales with the number of leaf as $O(|\mathcal{L}| \ln |\mathcal{L}|)$ in fixed DAGs (see Section.\ref{BAIMCDSkauf} in \textit{supplementary materials}), it is important to limit the number of added node $f(t)$ such that $f(t) \ln f(t) \ll t$, otherwise there is a risk that the algorithm never stops until all the leaf $\ell \in \mathcal{L}$ has been added. In particular, the usual expansion rule in growing MCTS that expands the tree at each iteration (UCT) is not suitable for BAI-exMCDS. On the other hand, \texttt{BAIExpand} should keep expanding the DAG on a regular basis to focus the search on the most promising arms.

Regarding the addition rule, the choice of \texttt{BAIAdd} is crucial to prevent the algorithm from allocating unnecessary resources to bad leaves. To do so, \texttt{BAIAdd} should account for the information gained during the search (empirical mean, number of node visits or RAVE score [\cite{browne2012survey}]).\\

We introduce a positive expansion parameter $b < 1$ and denote $T_s(t)$ the number of visits of the node $s$ when going down the DAG after $t$ steps. Finally, we introduce $\mathcal{C}_{\mathcal{D}_t}(s)$ the child node set of $s$ defined in the SR-DAG $\mathcal{D}_t \subseteq \mathcal{D}$.\\

\paragraph{\texttt{BAIAdd}$(\mathcal{D}_t,\mathcal{D})$:} Return $\mathcal{D}_{t+1}$ that is added a node $s_{\text{new}} \in \mathcal{V} \setminus \mathcal{V}_t$
to $\mathcal{D}_t$ together with all the edges $(s_1,s_{\text{new}})$ for $s_1\in \mathcal{V}_t$, where $s_{\text{new}}$ is selected as follows:
\begin{align*}    
  \mathcal{V}_{\text{expand}}(\mathcal{D}_t,\mathcal{D}) = & \left\{ s \in \mathcal{V}_t \mid \mathcal{C}_{\mathcal{D}_t}(s) \neq \mathcal{C}(s) \right\} \text{ : set of node that can be expanded},\\
  d(s) \phantom{=}& \text{ : depth of the node $s$ in the SR-DAG $\mathcal{D}$},\\
  I_s = & \frac{T_s}{\left|\mathcal{C}_{\mathcal{D}_t}(s)\right| + 1} \times \frac{1}{d(s)} \text{ : index introduced for each node $s \in \mathcal{V}_{\text{expand}}(\mathcal{D}_t,\mathcal{D})$},\\
  s_{\text{expand}} \gets & \max_{s \in \mathcal{V}_{\text{expand}}(\mathcal{D}_t,\mathcal{D})} I_s \text{ : node to be expanded}\\
  s_{\text{new}} \gets & \text{a node selected from }  \mathcal{C}(s_{\text{expand}}) \setminus \mathcal{C}_{\mathcal{D}_t}(s_{\text{expand}}) \text{ randomly}\\
  &\text{or use a scoring metric like RAVE score [\cite{browne2012survey}]}\\
  & \text{to further improve the expansion.}
\end{align*}      

Intuitively, a node should be expanded if it has been visited many times (hence the term $T_s$) or if it only has few children (hence the denominator). The term +1 is added to avoid giving an infinite index to temporary leaf nodes (that has no child by definition), and the factor $1/d$ is introduced to keep relatively balanced trees.\\

\paragraph{\texttt{BAIExpand}$(\mathcal{D}_t,t)$:} Return \texttt{True} if and only if $\lfloor (t+1)^{b} \rfloor - \lfloor t^b \rfloor = 1$ \\

This expansion rule is inspired from the discrete bandit policy introduced by \cite{gaudel2010feature} to deal with the large number of arms in UCT.\\

\subsection{Complexity term and sample complexity guarantees}

We define $\Delta_{\varepsilon,*} = \max (\Delta_*, \varepsilon)$, where $\Delta_* = V(s^{*}) - V(s^{*}_2)$ is the value difference between the best and the second best arm from the root node. 

\begin{theorem}\label{th:2} For $b \in \ ]0,1]$ and $\delta \leq \min (1, 0.1|\mathcal{L}_0|)$, under the condition that $\tau^b \gg |\mathcal{L}_0|$ and $\ln|\mathcal{L}_\tau| \gg \ln \ln \frac{1}{\Delta_{\varepsilon,*}^2}$, the number of leaf evaluation $\tau$ necessary to fulfill the stopping condition of LUCB-exMCDS is upper bounded by
\begin{equation}
\nonumber
    \tau \leq O \left( \left[ \delta^{\frac{b-1}{b}}  \exp \left( W_{-1}\left(
      \frac{\Delta_{\varepsilon,*}^2 (b-1)}{8 b \delta^{\frac{b-1}{b}}}
    \right) \right) \right] ^{\frac{1}{b-1}} \right),
\end{equation}
where $W_{-1}$ is the second real branch of the Lambert function [\cite{corless1996lambertw}].
\end{theorem}
The proof of Theorem~\ref{th:2} is directly related to the expansion policy \texttt{BAIExpand}, and is given in \textit{supplementary material} (Appendix~\ref{proof2}).\\

\hspace{-18pt}\textit{\textbf{Remark }The returned recommended arm by LUCB-exMCDS is $(\varepsilon,\delta)$-correct only within the leaves that has been evaluated $\mathcal{L}_\tau$ when the algorithm stops, there is no theoretical guarantee about the not yet expanded terminal leaves $\mathcal{L} \setminus (\mathcal{L}_\tau \cap \mathcal{L})$.}\\

Since Theorem~\ref{th:2} does not give an intuitively understandable result, we give an upper bound of the Lambert function [\cite{chatzigeorgiou2013bounds}]:
\begin{equation}
\nonumber
    \forall u > 0,\ \  W_{-1}\left(-e^{-u-1}\right) \leq -1 -\sqrt{2u} - \frac{2}{3}u
\end{equation}
Which gives 
\begin{equation}
    \nonumber
    \tau \leq O \left( \left[ \delta^{\frac{b-1}{b}}  \exp \left( -1 -\sqrt{2u} - \frac{2}{3}u \right) \right] ^{\frac{1}{b-1}} \right) \ \  \text{with} \ \ u = \ln \left( \frac{8 b \delta^{\frac{b-1}{b}}}{\Delta_{\varepsilon,*}^2 (1-b)}\right) -1
\end{equation}

Experimental validation of Theorem.\ref{th:2} is performed in \textit{supplementary materials} (Section \ref{experiment}). In particular, the advantage of considering SR-DAG architectures over trees, as well as a proof of concept for our expanding rule, is shown empirically on an hypothetical feature selection problem.

\newpage

\section{Best Leaf Identification in Monte Carlo DAG Search (BLI-MCDS)}

 The Best Leaf Identification problem consists of quickly and accurately finding the terminal leaf $\ell^* \in \mathcal{L}$ with the highest value,
\begin{equation}
  \nonumber
    \ell^{*} = \argmax_{\ell \in \mathcal{L}}V(\ell).
\end{equation}
In this section we propose an ($\varepsilon$, $\delta$)-PAC algorithm to identify the best leaf $\ell^*$ of a given DAG $\mathcal{D}$ by solving the BAI problem at each stage from the root until a terminal leaf is recommended. At each iteration, the algorithm BLI-MCDS (Best Leaf Identification in Monte Carlo DAG Search) starts from the root node and recursively select the recommended arm from the node $s$ if the stopping condition \texttt{BAIStop}($s$) is verified, but continue to look for the best child of $s$ if it is not. The algorithm stops when the BAI problem is solved at each stage of the DAG from the root node to a terminal leaf.

\begin{algorithm2e}
\renewcommand{\thealgocf}{BLI-MCDS}
\caption{Identify the best terminal leaf in a DAG $\mathcal{D}$}\label{alg:BLI-MCDS}

\DontPrintSemicolon
\SetKwFunction{FMain}{BLI-MCDS}
\SetKwFunction{Ft}{BLISelect}
\SetKwProg{Fn}{}{:}{}

\SetKwInOut{Input}{input}
\SetKwInOut{Output}{output}

\vspace{0.2cm}
  \Fn{\FMain{$\mathcal{D}_0$, $\mathcal{D}$, $\delta$, $\varepsilon$, $b$}}{

    $\mathcal{D}_t\gets \mathcal{D}_0$, $s_n \gets s_0$,  \ $ t \gets  0, \forall s \in \mathcal{D} \ t_s \gets 0$\\
    \While{\textbf{not} \emph{\texttt{BLIStop}$(s_n,\mathcal{D})$}}
    {
        \eIf{ $\emph{\texttt{BAIExpand}}(\mathcal{D}_t,t_{s_n})$}{
            $\mathcal{D}^{s_n}_t \gets$ subDAG of $\mathcal{D}_t$ under $s_n$\\
            $\mathcal{D}_{t+1}\gets \texttt{BAIAdd}(\mathcal{D}^{s_n}_t,\mathcal{D})$
        }{
            $\mathcal{D}_{t+1}\gets \mathcal{D}_t$
        }
    
        $s_n \gets \texttt{BLISelect}(s_0, \mathcal{D}_{t+1})$\\
        $R_{t+1} \gets \texttt{BAISelect}(s_n,\mathcal{D}_{t+1})$\\
        Sample the representative leaf $\ell_{R_{t+1}}$\\
        Update the confidence intervals for all the ancestors of $\ell_{R_{t+1}}(t)$.\\
        $t \gets t+1$\\
        $t_{s_n} \gets t_{s_n}+1$
    }
    \KwRet $s_n$
  }
\vspace{0.2cm}
  
  \SetKwProg{Pn}{}{:}{}
  \Pn{\Ft{$s,\mathcal{D}_t$}}{
    \uIf{$\emph{\texttt{BAIStop}}(s,\mathcal{D}_t)$}{
        $s_b \gets \texttt{BAIReco}(s,\mathcal{D}_t)$\;
        \KwRet $\texttt{BLISelect}(s_b,\mathcal{D}_t)$\;
    }
    \Else{
        \KwRet $s$\;
    }
  }
\vspace{0.2cm}

\texttt{BLIStop}($s,\mathcal{D}$): Return \texttt{True} if and only if $s \in \mathcal{L}$

\vspace{0.2cm}
    
\end{algorithm2e}
\begin{theorem}\label{th:3} For $\delta \leq \min (1, 0.1|\mathcal{L}|)$, the returned recommended leaf $\ell^*$ by \emph{\texttt{BLI-MCDS}} with an \emph{LUCB} BAI policy is ($\varepsilon$, $\delta$)-correct within all the leaf $\mathcal{L}_\tau$ that has been expanded at the end of the search.
\end{theorem}
The proof relies on the definition of the confidence intervals in the DAGs as well as the stopping rule $\texttt{BLIStop}$ and is given in \textit{supplementary materials} \ref{proof3}.\\

As for the expansion policy, because the best leaf is ($\varepsilon$, $\delta$)-correct only within the leaf that has been expanded at the end of the search, the way the DAG is expanded during the search is a major concern regarding the returned best leaf $\ell^*$. In particular, the expansion rules of LUCB-exMCDS introduced in section \ref{expansion_rule} are not optimal for BLI, because they were designed to solve a single BAI problem, and not multiple ones at the same time. One reason is that \texttt{BAIAdd} tends to expand nodes at low depth to keep balanced DAGs, and thus does not explore deep enough when BLI-MCDS is solving a BAI problem at deeper nodes. \texttt{BAIExpand} and \texttt{BAIAdd} are thus adapted to the BLI problem as follow:

\begin{itemize}

    \item $\texttt{BAIExpand}(\mathcal{D}_t,t)$ becomes $\texttt{BAIExpand}(\mathcal{D}_t,t_{s_n})$ where $t_{s_n}$ is the number of samples involved to solve the BAI problem from node $s_n$, so the expansion is \textit{reset} each time a new BAI problem is considered.
    
    \item $\texttt{BAIAdd}(\mathcal{D}_t,\mathcal{D})$ becomes $\texttt{BAIAdd}(\mathcal{D}^{s_n}_t,\mathcal{D})$ where $\mathcal{D}^{s_n}_t$ is the subDAG of $\mathcal{D}_t$ under $s_n$: the goal is to focus on the BAI problem that is currently being solved and thus only expand the nodes under it.
    
\end{itemize}

\section{Application : Feature Selection as a Best Leaf Identification problem}

Generally speaking, the gain of considering DAGs over trees in Monte Carlo Search for a given game mainly depends on the number of transpositions in its corresponding game state space. More transpositions results in a higher degree of redundancy in the tree architecture, and thus requires more episodes to get accurate estimates of action values. As a result, feature  selection  problems  provide  a  good example of the DAG's benefit over trees due to the very high degree of transposition in feature graphs.

\subsection{The game state space of Feature Selection}

Following in the footsteps of \cite{gaudel2010feature} who formalized Feature Selection as a reinforcement learning problem, we consider in this section Feature Selection as a Best Leaf Identification (BLI) problem. Given a feature set $\mathcal{F}$, we define a graph $\mathcal{D}=(\mathcal{V},\mathcal{E})$ for which each node $s_F \in \mathcal{V}$ corresponds to a feature subset $F\subset \mathcal{F}$, the number of distinct nodes in the graph is then $|P(\mathcal{F})| = 2 ^{|\mathcal{F}|}$. For a node $F$, we define the child nodes set $\mathcal{C}(s_F)$ and the parent node set $\mathcal{P}(s_F)$ as
\begin{equation}
\nonumber
\begin{split}
    &\mathcal{C}(s_F) = \left\{ s_{F_c} \in \mathcal{V} \mid \exists f \in \mathcal{F} \setminus F, \hspace{1mm} F_c = F \cup \{f\}  \right\},\\
    &\mathcal{P}(s_F) = \left\{ s_{F_p} \in \mathcal{V} \mid \exists f \in F, \hspace{1mm} F_p = F \setminus \{f\}  \right\}.\\
\end{split}
\end{equation}
%Each node then has $|\mathcal{C}(s_F)| = |\mathcal{F}|-|F|$ children and $|\mathcal{P}(s_F)| = |F|$ parents.

\begin{figure}[h!t]
    \centering
    \includegraphics[width=0.4\columnwidth]{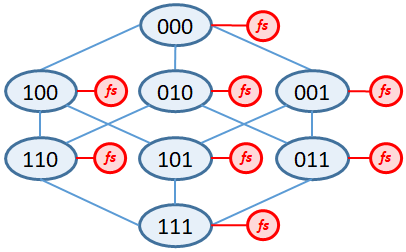}
    \caption{DAG architecture of a feature selection problem with a feature set of cardinal $|\mathcal{F}|=3$. The DAG contains $2^{3+1}=16$ distinct nodes, including  $2^{3}=8$ terminal leaves (shown in red). }
    \label{fig:fsfs}
\end{figure}

To allow the search to stop at the current node rather than adding new features to the subset, we also consider for each node an additional virtual stopping feature $f_s$, which always leads to a terminal leaf (Figure.\ref{fig:fsfs}). A leaf $\ell \in \mathcal{L}$ is said to be terminal if and only if its corresponding feature set contains the stopping feature. This multiply the number of nodes in the DAG by two and as a result we have $2^{|\mathcal{F}|+1}$ nodes in the DAG.

\subsection{Feature subset evaluation}

Let $\mathbb{X} = (x_i,y_i)$ be a training set containing $n$ labeled examples and $\mathbb{V}$ be a small subsample of $\mathbb{X}$ containing $m$ elements. For a given feature set $F$ associated to a leaf node $\ell_F$, the oracles are defined as follow:

\begin{itemize}
    \item The \textbf{Terminal Oracle} evaluates $F$ by computing the Area under the ROC Curve (AUC) with a $k$-NN classifier trained on the full dataset so that the value of the feature set is fully known after a single call and the upper bound and lower bound are directly set to that value. For a more efficient computational cost, a KD-trees algorithm is used: $\mu_F$ = \texttt{Evaluation}($F,n,\mathbb{X}$)
    
    \item The \textbf{Intermediate Oracle} evaluates $F$ by uniformly selecting features until the stopping feature is selected. At each round in the random phase, a new feature $f$ is added to $F$ ($F \gets F \cup \{ f \}$) whith the stopping feature being selected with a probability of $1-q^{|F|}$, where $q \in [0, 1]$ is a parameter of the algorithm. Once the stopping feature is selected, the obtained feature set $F^*$ is evaluated with a $k$-NN classifier trained a small subsample of the dataset to reduce computational cost, this evaluation is the same as the one introduced by \cite{gaudel2010feature}:\\
    $^*\mu_F$ = \textit{average}$\left\{\texttt{Evaluation}(F^*,m,\mathbb{X})\right\}$
\end{itemize}

\begin{algorithm2e}
\renewcommand{\thealgocf}{Evaluation}
\caption{Evaluate a feature subset with a $k$-NN classifier}\label{alg:evaluation}
\SetKwInOut{Input}{input}
\SetKwInOut{Output}{output}
    \Input{feature subset $F$, integer $m$, training set $\mathbb{X}$}
    \Output{score $V$}
    Compute $\mathbb{V}$ by uniformly selecting $m$ examples in the training set $\mathbb{X}$\\
    Train a $k$-NN classifier and for each example $x$ in $\mathbb{V}$, count for the number of positively labeled examples among its neighbours in $\mathbb{X}$ with features in $F$, $s_F(x) = \left|\left\{x′ \in \mathcal{N}_{F,k}(x), \hspace{2mm} y′ > 0\right\}\right|$\\
    Compute the Area Under the ROC curve as $V = \frac{\left| \left\{ (x,x') \ \in \mathbb{V}^2 , \hspace{2mm} s_F(x) < s_F (x') \hspace{1mm}, \hspace{2mm} y < y' \right\}  \right|}{ \left| \left\{ (x,x') \ \in \mathbb{V}^2 , \hspace{2mm} y < y' \right\}  \right|}$\\
    \textbf{return} $V$
%\end{algorithmic}
\end{algorithm2e}

\subsection{DAG Expansion and RAVE score}
The selection of new nodes can benefit from any knowledge gained within the search. \cite{gaudel2010feature} proposed to define a RApid Value Estimation (RAVE) score for the features, that are used to focus the search and avoid a (hopeless) uniform exploration of the feature space. The global and local RAVE scores of a feature $f$ are defined as follow:
\begin{equation}
\nonumber
    \begin{split}
        &\text{g-RAVE}_f = \textit{average}\{ V(F), \hspace{1mm} f \in F \}\\
        &\ell\text{-RAVE}_{F,f} = \textit{average}\{ V(F_t),\ F \subset F_t, \ f \in F_t \}
    \end{split}
\end{equation}
While g-RAVE provides a good global indication on feature relevance, it cannot account for redundancy relatively to the current node, thus it also makes sense to consider the feature conditionally to those selected within the current node, yielding to the $\ell$-RAVE factor.\\

Regarding the expansion policy, the new feature $f$ to be added by \texttt{BAIAdd} to the node $F$ is the one maximizing its RAVE score, that we calculate for each features by considering both g-RAVE and $\ell$-RAVE with the following expression:

\begin{equation}
\nonumber
    \begin{split}
        &\text{RAVE}_{F,f} \ = \ (1 - \beta_{F,f}) \ .\   \ell\text{-RAVE}_{F,f} \  + \ \beta_{F,f} \ .\ \text{g-RAVE}_f \hspace{3cm}\\
        &\text{with} \ \ \beta_{F,f} = \frac{c_l}{c_l + t_{F,f}}  
    \end{split}
\end{equation}

$t_{F,f}$ is the number of iterations involved in $\ell\text{-RAVE}_{F,f}$ computation, and $c_l$ is a fixed parameter to tune the importance of $\ell\text{-RAVE}$ relatively to $g\text{-RAVE}$. When the information about $\ell\text{-RAVE}_{F,f}$ is inaccurate due to a low number of computation, we consider the g-RAVE score instead, but gradually account for $\ell\text{-RAVE}_{F,f}$ as its value gets more reliable. The RAVE score of the stopping feature $f_s$ is set to infinity to ensure that the terminal child leaf is always the first to be expanded.

\begin{table*}[ht]
\label{table:results}
\centering
  \caption{Sample complexities, recommended leaf value ($V$) and number of features $f$ selected by BLI-MCDS on benchmark data set with $b=0.3$, $\varepsilon = 0.005$ and $\delta = 0.1$, results are averaged over 15 times. A comparison with the FUSE algorithm is given (when same number of iteration is applied). We recall $\beta_1(\delta,N) = \ln \frac{|\mathcal{L}|}{\delta} + \ln (\ln(N) + 1 )$, 
$\beta_2(\delta,N) = \ln \left(\frac{\ln (eN)}{\delta}\right)$
}
\begin{tabular}{|c|c|c||c|c|c||c|c|}
  \hline
  \multirow{2}*{ Data set} & \multirow{2}*{ Properties} & \multirow{2}*{$\beta(\delta,N)$} & \multirow{2}*{\#sample} & \multirow{2}*{$V(\ell^*)$} & \multirow{2}*{\#$f$} & \multirow{2}*{$V(\ell_{\text{FUSE}})$} &       \multirow{2}*{\#$f_{\text{FUSE}}$}\\
  & & & & & & &\\
  \hline
  \multirow{2}*{Linear} & 300 examples & $\beta_1$ & 5,032,404 & 0.9943 & 3.00 & 0.9873 & 2.33\\
  & 30 features &$\beta_2$ & 440,240 & 0.9943 & 3.00 & 0.9872 & 2.40\\
  \hline
  \multirow{2}*{Madelon} & 2000 examples & \multirow{2}*{$\beta_2$} & \multirow{2}*{2,496,391} & \multirow{2}*{0.9459} & \multirow{2}*{7.20} & \multirow{2}*{0.9473} & \multirow{2}*{6.33}\\
  &  500 features & & & & & &\\
  \hline
\end{tabular}
\end{table*}

\subsection{Experiments}

To find the best feature subset, \cite{gaudel2010feature} introduced the algorithm FUSE, that starts from the empty feature subset and relies on UCT to identify the best leaf $\ell^*$, which is taken at the end of the search as the path with the highest average at each step from the root node. In their study, they show that FUSE can compete with the state of the art feature selection algorithms (Correlation-based Feature Selection, Random-Forest based Gini score). Comparing our ($\delta$, $\varepsilon$)-PAC \textit{setting} with FUSE is not obvious as it would require to define a suitable stopping rule for UCT, but it is possible to compare the performances by measuring the number of iterations required to fulfill the stopping condition of BLI-MCDS and then run FUSE with the same number of iterations.

In our experiment, BLI-MCDS was run with parameter $b=0.3$, that is a good compromise between the relatively high number of leaf to be explored and the resulting increased computational complexity (see section \ref{BAIex_complexity} in \textit{supplementary materials}). Each time the BAI problem was solved at a node, The DAG below the optimal arm was instantly initialized at depth 2 with 7 features that maximize their RAVE score previously computed. Two exploration functions, theoretically guaranteed $\beta_1$ and empirically optimized $\beta_2$ previously discussed in section \ref{exploration_beta}, were used. $\beta_2$ is more practical because it is independent to the number of leaves in the DAG and thus does require to update all the nodes after each expansion. Regarding the evaluation oracles, we set $m = 50$, $k = 5$ and $q = 0.9$, whose values were discussed by \cite{gaudel2010feature}.

Two dataset were used : The first one is a simple linear artificial data set with 30 features, containing 3 features $x,y,z$, plus 7 redundant and 20 randomly generated features. To build the data set, 300 examples were generated by randomly selecting uniformly $(x,y,z)$ in $[0,1]^3$, and a linear classification function was used: $f(x,y,z) = (0.1x -0.8y + 0.6z > 0)?$ to calculate the labels. The second one is the Madelon dataset [\cite{guyon2003design}], which was designed for the NIPS 2003 feature selection challenge [\cite{guyon2005result}]. It is an artificial 500-features dataset, where the target concept is set to five relevant features. The other 495 features involve 15 redundant features, built as linear combinations of the relevant ones, and the remaining features are irrelevant. The data set is provided with 2000 examples. Each algorithm was run independently 15 times\footnote{The \texttt{C++} implementation for Feature Selection with FUSE and BLI-MCDS is available on Github at \href{https://github.com/Aurelien-Pelissier/Feature-Selection-as-Reinforcement-Learning}{https://github.com}}, and the results are summarized Table \ref{table:results}.\\

While our algorithm shares some similarity with UCT by going down the DAG choosing the child that has the highest Upper Confidence Bound, it relies on confidence intervals based on descendant confidence bounds, whereas UCT uses confidence intervals based on the number of visits of the parent node. A well-known problem of UCT is that it sometimes take an impractically long time to find the best leaf as the asymptotic regret of $O(\log(n))$ might involve a very long transitory phase [\cite{coquelin2007bandit}]. This is illustrated by the result of the first dataset, where FUSE failed to find the best leaf at depth 3 because its value is \textit{drawn} by the average of other leaves.

On the other hand, one weakness of BLI-MCDS is that it explores a relatively low number of leaves compare to FUSE, and thus is heavily dependant on the expansion policy \texttt{BAIAdd} and RAVE score. As a result, FUSE seems to slightly outperform BLI-MCDS when the branching factor and the target number of relevant features is relatively large (Madelon data set).

\section{Summary and Discussions}

We introduced a novel algorithm for the Feature Selection problem with theoretical guaranties by representing the feature set space as a SR-DAG and by recursively solving the BAI problem at each stage from the root to a terminal leaf to identify the best leaf node. We focused on the sample complexity of Monte-Carlo DAG Search methods, about which very little was known, and we developed a new BAI algorithm in expanding single rooted DAG, that has a considerable potential in reinforcement learning problems. 

While our BLI algorithm shares some similarity with FUSE by looking for the best leaf in the feature DAG, it relies on very different selection and expansion policies, that we highlighted in our experiments. Furthermore, \cite{gaudel2010feature} originally introduced FUSE with an \textit{update-descent} backpropagation policy (updating only the traversed edges during the UCT phase), whereas BLI-MCDS is backpropagating the confidence intervals with an \textit{update-all} policy (updating every ancestor nodes), which is another benefit of BLI-MCDS over FUSE.\\

\hspace{-18pt}\textbf{Future directions }$\ $ As for further research, improvements on the expansion policy for BAI-exMCDS and BLI-MCDS would be beneficial as it would focus the search on more promising nodes. In particular, defining an expansion policy that provides ($\varepsilon$, $\delta$)-correctness on $\mathcal{L}$ rather than $\mathcal{L}_{\tau}$ would be a significant enhancement. It would also be interesting to adapt FUSE with UCD [\cite{saffidine2012ucd}] (that is optimized for DAGs) to further improve the performance of FUSE and gives a more fair comparison to BLI-MCDS. One could also consider accounting for cycles in the game state space and further generalize BAI-exMCDS to connected graph. The question is related to the Graph History Interaction (GHI) problem for which a general solution was proposed by \cite{kishimoto2004general}.

\section*{Acknowledgments}
Aur\'elien P\'elissier acknowledges the financial support from the \'Ecole Normale Sup\'erieure (ENS) Paris-Saclay and Hokkaido University.

\bibliography{sdm2019}

\appendix

\newpage

\begin{center}{\Huge Supplementary materials}\end{center}

\vspace{0.2cm}

\section{Complexity term and sample complexity guarantees of LUCB-exMCDS without expansion}

\label{BAIMCDSkauf}

In this section we assume that $b=0$ and thus $\forall t \in \mathbb{N}, \mathcal{D}_t = \mathcal{D}$. For further analysis of the LUCB-exMCDS algorithm, we introduce the value difference $\Delta V(s)$ of a node $s$ with its parents as 
\begin{equation}
\nonumber
    \Delta V (s) = \max_{p \in \mathcal{P}(s)} \left|V(s) - V(p)\right|.
\end{equation}
We define the second best arm at the root node:
% (we do not consider the arms that have the same representative leaf as the best arm $s^*$)
\begin{equation}
\nonumber
    s^{*}_2 = \argmax_{s \in \mathcal{C}(s_0),V(s)\neq V(s_0)} V(s).
\end{equation}
Let $\text{Anc}(\ell)$ be the node set of the ancestors of $\ell$, the complexity term $H_\varepsilon$ is defined as
\begin{equation}
\nonumber
    \begin{split}
        &H_\varepsilon = \mathlarger{\mathlarger{\sum}}_{\ell \in \mathcal{L}} \ \dfrac{1}{\max (\Delta_\ell^2, \Delta_*^2, \varepsilon^2) } \\
        &\text{where} \ \ \ \begin{array}{cc} \ \Delta_* = V(s^{*}) - V(s^{*}_2),\ \ \ \ \ \ \ \ \ \ & \\ \ \Delta_\ell = \max_{s \in \text{Anc}(\ell)  \backslash \{s_0\}}\Delta V (s).& \end{array}
    \end{split}
\end{equation}
The meaning of these squared terms in the denominator is the following:\\
\begin{tabular}{lp{13cm}}
$\frac{1}{\varepsilon^2}$: &  \textit{number of samples needed for $\hat{\mu}_{\ell}$ reaching the required precision},\\
$\frac{1}{\Delta_*^2}$: & \textit{number of samples needed for distinguishing the best from the second best (this happens for leaves below the optimal arm)},\\
  $\frac{1}{\Delta_\ell^2}$:&  \textit{number of samples needed for noticing existence of a leaf better than $\ell$
    (if you can find a path where $\ell$ is a bad leaf, then you can prune it from all paths.)}
\end{tabular}

\begin{theorem}\label{th:1} For $b=0$ and $\delta \leq \min (1, 0.1|\mathcal{L}|)$, the number of leaf evaluation $\tau$ necessary to fulfill the stopping condition of LUCB-MCDS is upper bounded by
\begin{equation}
\nonumber
    \begin{split}
        &\tau \leq8 H_\varepsilon \ln \frac{|\mathcal{L}|}{\delta}
        + \mathlarger{\mathlarger{\sum}}_{\ell \in \mathcal{L}} \hspace{2pt} \frac{16}{\overline{\Delta}_{\ell,\varepsilon}^2} \ln \ln \frac{1}{\overline{\Delta}_{\ell,\varepsilon}^2} +  o \left(\ln \frac{|\mathcal{L}|}{\delta} \right)\\
        & \text{with} \ \ \ \overline{\Delta}_{\ell,\varepsilon} = \max (\Delta_\ell, \Delta_*, \varepsilon).   
    \end{split}
\end{equation}
\end{theorem}

\hspace{-18pt}\textit{\textbf{Remark }Although this formula is very similar to the one described by~\cite{kaufmann2017monte} and \cite{huang2017structured}, one should note the difference in the number of ancestors involved in $\Delta_\ell$, which can be exponentially larger for DAGs with relatively high number of edges.}\\

The proof is straight forward from the previous results of \cite{kaufmann2017monte} and the different choice of $b_t$ only slightly affect the proof, details are given in Appendix~\ref{proof1}. An interesting result is that the first term scales with the number of leaf as $O(|\mathcal{L}| \ln |\mathcal{L}|)$. Regarding the second term, \cite{jamieson2014lil} proved that it is unavoidable when the gaps between arms are small, but it becomes negligible when
$$\ln|\mathcal{L}| \gg \ln \ln \dfrac{1}{\max (\Delta_*^2, \varepsilon^2)}.$$
One should note that Theorem~\ref{th:1} holds for any SR-DAG configuration, and that the sample complexity mainly depends on the leaf nodes $\mathcal{L}$ rather that the tree architecture. In particular, the sample complexity is not significantly affected by unbalanced DAGs.

\section{Experimental Validation of Theorem.\ref{th:1} and Theorem.\ref{th:2}}

\label{experiment}

In this section, we consider an hypothetical feature selection problem as a benchmark for our experiment, because feature selection problems provide a perfect example of the DAG's benefit over trees due to the high degree of transposition in feature graphs.

\subsection{The feature graph}
Given a feature set $\mathcal{F}$, we define a graph $\mathcal{D}=(\mathcal{V},\mathcal{E})$ for which each node $s_F \in \mathcal{V}$ corresponds to a feature subset $F\subset \mathcal{F}$, the number of distinct nodes in the graph is then $|P(\mathcal{F})| = 2 ^{|\mathcal{F}|}$. For a node $F$, we define the child nodes set $\mathcal{C}(s_F)$ as
\begin{equation}
\nonumber
    \mathcal{C}(s_F) = \left\{ s_{F_c} \in \mathcal{V} , \hspace{1mm} \exists f \in \mathcal{F} \setminus F, \hspace{1mm} F_c = F \cup \{f\}  \right\}.\\
\end{equation}
Each node then has $|\mathcal{C}(s_F)| = |\mathcal{F}|-|F|$ children.\\

\setlength{\columnsep}{25pt}
\begin{multicols}{2}

\noindent
\textbf{SR-DAG:}\\
We define the parent node set $\mathcal{P}(F)$ as
\begin{equation}
\nonumber
    \mathcal{P}(s_F) = \left\{ s_{F_p} \in \mathcal{V} , \hspace{1mm} \exists f \in F, \hspace{1mm} F_p = F \setminus \{f\}  \right\}\\
\end{equation}
Each node then has $|\mathcal{P}(s_F)| = |F|$ parents.\\

\noindent
\textbf{Tree:}\\
Because the number of parents allowed in trees is fixed to one, it is necessary to introduce redundant nodes to conserve the relationship between each feature subset.

\end{multicols}

\begin{figure}[h!t]
    \centering
    \includegraphics[width=1\columnwidth]{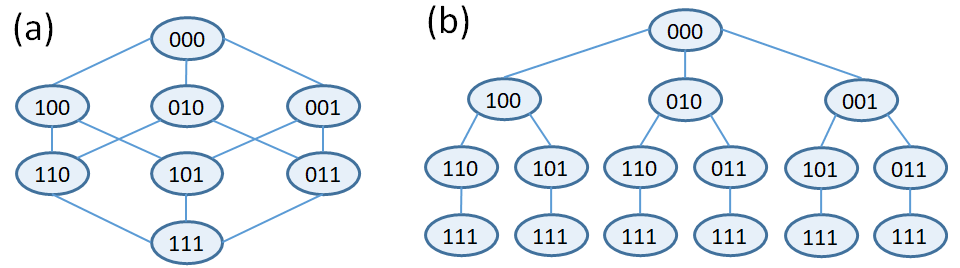}
    \caption{(a) Lattice and (b) tree architecture of a feature selection problem with a feature set of cardinal $|\mathcal{F}| = 3$. The DAG contains $2^3 = 8$ distinct nodes, whereas the tree has $16$ node.}
    \label{fig:DAG}
\end{figure}

Figure \ref{fig:DAG} clearly highlights the potential of considering a DAG over a tree. Intuitively, tree architectures result in a high degree of redundancy, and consequently BAI requires more episodes to get accurate estimates of action values.

\subsection{Feature subset evaluation}
For a simplified approach, each feature $f \in \mathcal{F}$ is associated to a score and each feature subset $F$ is then evaluated with
\begin{equation}
\nonumber
    \begin{split}
        &\mu_F = \text{sig} \left( \sum_{f \in F} \text{score}(f) \right) \\
        &\text{where} \ \ \ \ \ \text{sig}(x) = \frac{1}{1+ e^{-x}} \ \ \text{is the sigmoid function}
    \end{split}
\end{equation}
The sigmoid function ensures that the value $\mu_F$ is always within [0,1]. For our experiment, we focus on binary oracle $\mathcal{O}_\ell$ with a Bernoulli distribution whose mean is $\mu_F$.

\subsection{Benefit of fixed DAGs over fixes trees}
We first consider an hypothetical simplified feature selection problem for which we only evaluate nodes at a fixed depth $d_\mathcal{L}$.
\begin{equation}
\nonumber
    \mathcal{L} = \left\{ s_F \in \mathcal{V}, \ |F| = d_\mathcal{L} \right\}.
\end{equation}
The BAI problem at the root $s_0 = s_\emptyset$ corresponds to finding the best feature to be added, (note that there can be multiple best arms).
\begin{equation}
\nonumber
    s^* = \argmax_{f \in \mathcal{F}} V(s_{\{ f \}}).
\end{equation}

We implement both LUCB-MCDS and LUCB-MCTS with their corresponding graph being respectively a lattice and a redundant tree structure (described by Figure~\ref{fig:DAG}), and measure how many iterations is required to fulfill their stopping condition. For our experiment, each feature $f_i \in \{f_1 ... f_6\}$ is associated to a $Score(f_i) \in \{-0.3, 0, 0.03, 0.3, 0.4, 0.5\}$. We use a leaf depth of $d_\mathcal{L} = 3$, and a feature set of size $|\mathcal{F}| = 6$, leading to 20 leaf nodes $\left(_6C_3\right)$ in the DAG and 120 leaves in the tree $\left(\frac{6!}{(6-3)!}\right)$. As a result, the best leaf is $l^* = s_{\{ f_4, f_5, f_6  \}}$ with a value of $\mu_{l^*} = 0.77$ and the second best leaf is $l^{*}_2 = s_{\{f_3, f_5, f_6 \}}$ with a value of $\mu_{l^{*}_2} = 0.72$. In this particular problem, the best arms  are either $s_{\{f_4\}}$, $s_{\{f_5\}}$ or $s_{\{f_6\}}$, and the second best arm is $s_{\{f_3\}}$, leading to $\Delta_* = 0.05$. 

Each algorithm are run independently 10,000 times with the two exploration functions, theoretically guaranteed $\beta_1$ and empirically optimized $\beta_2$ previously discussed in section \ref{exploration_eq}.

\begin{table*}[ht]
\centering
  \caption{Sample complexities and correct recommendation rates of LUCB-MCTS and LUCB-MCDS for our feature selection problem with $\varepsilon = 0$ and $\delta = 0.1$. The theoretical sample complexity upper bound $\tau_\text{ub} =  8 H_\varepsilon \ln \frac{|\mathcal{L}|}{\delta}$ is also given for comparison (other terms of Theorem \ref{th:1} are negligibles in our experiment). We recall $\beta_1(\delta,N) = \ln \frac{|\mathcal{L}|}{\delta} + \ln (\ln(N) + 1 )$, 
$\beta_2(\delta,N) = \ln \left(\frac{\ln (eN)}{\delta}\right)$
}
\begin{tabular}{|c|l|c|r|r|c|}
  \hline
  \multirow{2}*{$\beta(\delta,N)$} & \multirow{2}*{method} & \multirow{2}*{$\tau_\text{ub}$} & \multirow{2}*{\#sample} & \multirow{2}*{\#(node update)} & root-child recommendation \\
  \cline{6-6}
  & & & & & $\frac{\#(s_{\{f_1\}},s_{\{f_2\}},s_{\{f_3\}},s_{\{f_4\}},s_{\{f_5\}},s_{\{f_6\}})}{10,000}$\\
  \hline
  \multirow{2}*{$\beta_1$} & MCTS & 498,521 & 478,142 & 1,912,568 & (0,0,0,0.33,0.33,0.33)\\
  & MCDS & 96,520 & 92,446 & 739,568 & (0,0,0,0,0,1.0)\\
\hline
\multirow{2}*{$\beta_2$} & MCTS & | & 27,874 & 111,496 & (0,0,0,0.33,0.33,0.33)\\
& MCDS & | &  17,495 & 139,960 & (0,0,0,0,0,1.0)\\
\hline
\end{tabular}
\end{table*}

While the error rate (0\%) remains unchanged in all cases, the number of steps involved in BAI-MCTS and BAI-MCDS clearly emphasizes the benefit of DAGs over trees, and the sample complexity is shown to be improved by a factor of $\sim$5. Note that the ratio is consistent with the theoretical upper bounds from Theorem~\ref{th:1}, and that improvements are expected to be even more significant for deeper leaves as the redundancy in tree structures becomes larger.

Although the use of $\beta_2$ is not theoretically guaranteed, the empirical exploration clearly outclass $\beta_1$ by a factor of $\sim$10. One can also note the less significant but still noticeable improvement of MCDS over MCTS when using $\beta_2$.

\subsection{Sample complexity verification of expanding DAG}
\label{BAIex_complexity}

In the case of expanding DAG, the initial number of leaf should be at least two, otherwise the stopping condition is fulfilled at the first iteration. For our experiment, we consider an initial DAG built with $|\mathcal{F}| = 15$ features at depth one, resulting in $|\mathcal{L}_0| = 15$ initial leaves. Since we are only interested in sample complexity verification, we associate to each node the same score for simplicity $(\forall s \in \mathcal{D}, V(s) = 0.5)$, the algorithm will then stops when it reaches the required precision $\varepsilon$ for each arm.

We run LUCB-MCDS with different expansion parameter $b$ and compare the number of iterations to the calculated upper bound $\tau_{\max}$ from Theorem~\ref{th:2}. The results for $\delta = 0.1$ and $\varepsilon = 0.05$ are plotted Figure.\ref{fig:expansion}:

\begin{figure}[h!t]
    \centering
    \includegraphics[width=0.5\columnwidth]{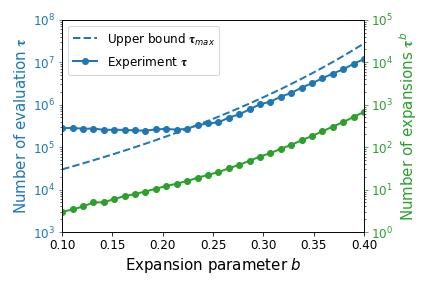}
    \caption{Number of iteration before LUCB-exMCDS stops as a function of the expansion parameter $b$, each averaged over 10 times.}
    \label{fig:expansion}
\end{figure}

The necessity of our assumption $\tau^b \gg |\mathcal{L}_0|$ now become very clear. With low expansion parameters, only few leaves are added to the DAG and thus the inequality does not hold (we recall that $|\mathcal{L}_0| = 15$). For $b>0.25$, the number of expansion becomes large enough (around 20 expansions) so that Theorem~\ref{th:2} becomes valid in our experiment.

One remark is that an algorithm which expands several dozens of nodes before stopping would not be sufficient in most applications, as it is limiting the search to only few positions. Regarding values of $b$ higher than 0.4, the number of samples to stop the algorithm becomes relatively large and thus less practical ($10^7$ in our experiment).

\section{Proof of Theorem~\ref{th:1}}

The generalization of BAI's sample complexity from trees to DAG is straight forward, but the change in the way $b_t$ is picked in \texttt{BAISelect} and  \texttt{BAIReco} leads to some modifications in the proof of \cite{kaufmann2017monte}. Their Lemma 13 is specific to UGapE-MCTS, and thus requires some adjustments to be applied to LUCB-MCDS. One should note that our choice of $b_t$ is the same as the LUCB-micro algorithm described by \cite{huang2017structured} and that they have obtained an upper bound very similar to ours.\\

\hspace{-18pt}\textbf{Lemma 12.} is a consequence of the definition of
the exploration procedure and does not depend on the way the optimal arm $b_t$ is picked.\\
\textit{If} $\ell$ \textit{is selected at round} $t + 1$, \textit{then}

\begin{equation}
\nonumber
 \sqrt{\frac{2 \beta(N_\ell(t),\delta)}{N_\ell(t)}} \geq \max_{k = 2...D} \left|V(s_k^i) - V(s_{k-1}^i)\right|
\end{equation}\\

\hspace{-18pt}\textbf{Lemma 13.} partially relies on Lemma.15 and exploits the mechanism of LUCB-exMCDS to relate $N_\ell$
to $\Delta_{*}$ and $\varepsilon$.\\
\textit{If} $\ell$ \textit{is selected at round} $t + 1$, \textit{then}

\begin{equation}
\nonumber
 \sqrt{\frac{8 \beta(N_\ell(t),\delta)}{N_\ell(t)}} \geq \max \left(V(s_0) - V(s_1^i),\ \varepsilon, \ \Delta_* \right)
\end{equation}\\

\subsection{Lemma 15}
The only part where the definition of $b_t$ has an impact on the proof is in their Lemma.15, that is straight forward to prove with our choice of $b_t$:\\

\hspace{-18pt}\textbf{Lemma 15.} $\forall t \in \mathbb{N}^*$\textit{, the following holds}
\begin{equation}
\nonumber
    \begin{split}
        \textit{if} \  R_{t+1} &= b_t \ \ \textit{then} \ \ U_{c_t}(t) \leq U_{b_t}(t)\\
        \textit{if} \  R_{t+1} &= c_t \ \ \textit{then} \ \ L_{c_t}(t) \leq L_{b_t}(t)\\
    \end{split}
\end{equation}

\hspace{-18pt}\textit{Proof if $c_t$ is selected}.\\
$L_{c_t}(t) \leq L_{b_t}(t)$ is a consequence of the definition of $b_t$.\\

\hspace{-18pt}\textit{Proof if $b_t$ is selected}.\\
Assume that $U_{c_t}(t) > U_{b_t}(t)$, as the confidence interval on $V(c_t)$ is larger than the confidence intervals on $V(b_t)$ (because $c_t$ is selected), this also yields to $L_{c_t}(t) > L_{b_t}(t)$, which contradict the definition of $b_t$. Thus, we proved by contradiction that $U_{c_t}(t) \leq U_{b_t}(t)$. \begin{flushright} $\blacksquare$ \end{flushright}

\subsection{Proof of Theorem~\ref{th:1}}

\label{proof1}
Let $t \in \mathbb{N}$ and ($s_0, s_1^i, ..., s_{D-1}^i, s_D$) be a path $\mathcal{M}_i$ from the root $s_0$ down to a leaf $\ell = s_D$\\

\hspace{-18pt}We define 
\begin{equation}
\nonumber
    \Delta_\ell^i = \max \left (V(s_0) - V(s_1^i), \ \max_{k=2...D} \left|V(s_k^i) - V (s_{k-1}^i)\right| \right)
\end{equation}
Since Lemma 12 and Lemma 13 hold for any path $\mathcal{M}_i$ from $s_0$ down to $\ell$, we obtain
\begin{equation}
\label{bound}
     \sqrt{\frac{8 \beta(N_\ell(t),\delta)}{N_\ell(t)}} \geq \max \left(\Delta_\ell,\ \varepsilon, \ \Delta_* \right) \ \ \ \ \text{with} \ \ \ \ \Delta_\ell = \max_{\mathcal{M}_i} \Delta_\ell^i
\end{equation}
Finally, The derivation of Theorem~\ref{th:1} from eq.(\ref{exploration_eq}) and eq.(\ref{bound}) is exactly the same as the one presented by \cite{kaufmann2017monte} for their Theorem 3. \begin{flushright} $\blacksquare$ \end{flushright}

\section{Proof of Theorem~\ref{th:2}}
\label{proof2}
The proof relies on the expansion rule \texttt{BAIExpand}, that decreases the number of expansions as the number of iterations $t$ increases. Let $t>0$, due to the expansion policy, it is possible that some nodes $s$ that were initially leaves $s \in \mathcal{L}_0$ has been expanded and the information about their evaluation has been lost ($s \notin \mathcal{L}_t$). Let $\mathcal{L}^*_t = \mathcal{L}_0 \cap \mathcal{L}_1 \cap .. \cap \mathcal{L}_t$ be the set of all the nodes that have been a leaf at least for one iteration during the search after $t$ iterations.\\

We start with the following proposition that is a direct consequence of the expansion rule definition \texttt{BAIExpand}.
\begin{proposition}\label{pr:1} \textit{With the LUCB-exMCDS algorithm, the total number of leaves $|\mathcal{L}^*_t|$ explored after $t$ steps is upper bounded by}
\begin{equation}
\nonumber
    |\mathcal{L}_t^*| \leq |\mathcal{L}_0| + t^b
\end{equation}
\end{proposition}
\textbf{Proof.} From the definition of the expansion rule \texttt{BAIExpand}, the number of expansions in the DAG is $\lfloor t^b \rfloor \leq t^b$. One expansion explore one new leaf node (\texttt{BAIAdd}), and thus $|\mathcal{L}^*_{t+1}| = |\mathcal{L}^*_t| + 1$. A Recursive reasoning from $t=0$ conclude the proof. \begin{flushright} $\blacksquare$ \end{flushright}

\hspace{-18pt}\textbf{Proof of Theorem~\ref{th:2}.}

\hspace{-18pt}From Theorem~\ref{th:1} we can write the upper bound of the number of leaf evaluations $\tau$ that are necessary to fulfill the stopping condition of LUCB-exMCDS, we recall $\overline{\Delta}_{\ell,\varepsilon} = \max (\Delta_\ell, \Delta_*, \varepsilon)$,
\begin{equation}
\nonumber
    \tau \leq 8H_\varepsilon \ln \left(\frac{|\mathcal{L}^*_\tau|}{\delta}\right)  + \mathlarger{\mathlarger{\sum}}_{\ell \in \mathcal{L}_\tau} \hspace{2pt} \frac{16}{\overline{\Delta}_{\ell,\varepsilon}^2} \ln \ln \frac{1}{\overline{\Delta}_{\ell,\varepsilon}^2} + o \left( \ln \frac{|\mathcal{L}^*_\tau|}{\delta} \right)
\end{equation}
The first term scales with the number of leaf as $O\left(|\mathcal{L}^*_\tau|\ln |\mathcal{L}^*_\tau|\right)$ while second term as $O(|\mathcal{L}^*_\tau|)$, which becomes negligible when the number of leaf is large. In the following we assume that the gap between leaves $\overline{\Delta}_{\ell,\varepsilon}$ is large enough so that the first term becomes dominant ($\ln|\mathcal{L}^*_\tau| \gg \ln \ln \frac{1}{\overline{\Delta}_{\ell,\varepsilon}}$). To make the scaling with the number of leaf to appear more clearly, we rewrite the inequality
\begin{equation}
\nonumber
    \begin{split}
        &\tau \leq 8 A_{\mathcal{L}^*_\tau} \ |\mathcal{L}^*_\tau| \ \ln \left(\frac{|\mathcal{L}^*_\tau|}{\delta}\right) \\ 
        &\text{with} \ \ \ \ A_{\mathcal{L}^*_\tau} = \dfrac{1}{|\mathcal{L}^*_\tau|} \mathlarger{\mathlarger{\sum}}_{\ell \in \mathcal{L}^*_\tau} \ \dfrac{1}{\max (\Delta_\ell^2, \Delta_*^2, \varepsilon^2) }   
    \end{split}
\end{equation}
\begin{equation}
\nonumber
    \text{We have} \ \ 0 < A_{\mathcal{L}^*_\tau} \leq \dfrac{1}{\Delta_{\varepsilon,*}^2} \ \ \ \text{where} \ \ \ \Delta_{\varepsilon,*} = \max (\Delta_*, \varepsilon) \hspace{4.4cm}
\end{equation}
The total number of leaf evaluation $\tau$ is then upper bounded by
\begin{equation}
\nonumber
    \Delta_{\varepsilon,*}^2 \hspace{1pt} \tau \leq 8 \left( |\mathcal{L}_0| + \tau^b \right) \ln \left(\frac{|\mathcal{L}_0| + \tau^b}{\delta}\right)
\end{equation}
While this inequality does not have analytically solution, we can estimate the upper bound $\tau_\text{max}$ by assuming that $\tau_\text{max}^b \gg |\mathcal{L}_0|$,
\begin{equation}
\nonumber
    \Delta_{\varepsilon,*}^2 \tau_\text{max} = 8 \hspace{1pt} \tau_\text{max}^b \ln \left(\frac{\tau_\text{max}^b}{\delta}\right),
\end{equation}
Which leads to the solution
\begin{equation}
\nonumber
    \tau_\text{max} = \left[ \hspace{1pt}\delta^{\frac{b-1}{b}} \hspace{2pt} \exp \hspace{-3pt} \left( W\hspace{-3pt}\left(
    \frac{\Delta_{\varepsilon,*}^2 \hspace{1pt} (b-1)}{8 \hspace{1pt} b \hspace{1pt} \delta^{\frac{b-1}{b}}}
    \right) \right) \right] ^{\frac{1}{b-1}}
\end{equation}
Where $W$ is the Lambert function defined such as $x \ln (x) = y \Leftrightarrow x = e^{W(y)}$. Since $0<b<1$, the term inside $W$ is negative, and two solutions exists \cite{corless1996lambertw} for
\begin{equation}
\nonumber
    -\frac{1}{e} < \frac{\Delta_{\varepsilon,*}^2 \hspace{1pt} (b-1)}{8 \hspace{1pt} b \hspace{1pt} \delta^{\frac{b-1}{b}}} < 0
\end{equation}
The branch $W_0$ of the Lambert functions gives values close to zero, which is in contradiction with our assumption $\tau_\text{max}^b \gg |\mathcal{L}_0|$. Thus only the branch $W_{-1}$ remains, concluding the proof. \begin{flushright} $\blacksquare$ \end{flushright}

\section{Proof of Theorem~\ref{th:3}}

The proof relies on both the confidence interval definition in the DAG and our recommendation rule \texttt{BAIReco}. In particular, the choice of $b_t$ maximizing the lower bound is crucial for the ($\varepsilon$, $\delta$)-correctness of the recommended leaf. \\ 

\label{proof3}

    Assuming that the algorithm BLI-MCDS stops after $t = \tau$ iterations, let 
    $$\ell_b = \argmax_{\ell \in \mathcal{L}_\tau} \text{L}_\ell(\tau) \ \ \text{and} \ \ \ell_c = \argmax_{\ell \in \mathcal{L}_\tau \setminus \{\ell_b\}} \text{U}_\ell(\tau)$$ 
    be the candidate for the best leaf and the optimistic contender leaf respectively. Let $s_b \in \text{Anc}(\ell_b)$ and $s_c \in \text{Anc}(\ell_c)$ such that $s_b \neq s_c$ and $\mathcal{P}(s_b) \cap \mathcal{P}(s_c) \neq \varnothing$. Let $s_p \in \mathcal{P}(s_b) \cap \mathcal{P}(s_c)$ be one of their common parents.\\

    Let $s_0, s_1,... , s_D$ be a path from the root $s_0$ down to a leaf $\ell = s_D$. We remind the definition of the confidence intervals (Section.\ref{confidenceInterval}):
    \begin{equation}
    \nonumber
        \begin{split}
            &\text{L}_{s_{k−1}}(\tau) = \max_{s\in \mathcal{C}(s_{k−1})} \text{L}_s(\tau) \geq \text{L}_{s_{k}}(\tau),\\
            &\text{U}_{s_{k−1}}(\tau) = \max_{s\in \mathcal{C}(s_{k−1})} \text{U}_s(\tau) \geq \text{U}_{s_{k}}(\tau),\\
        \end{split}
    \end{equation}
    which directly gives $\text{U}_{s_c}(\tau) \geq \text{U}_{\ell_c}(\tau)$.\\
    
    By definition of the stopping rule $\texttt{BLIStop}$, the BAI problem is solved at each node from the root $s_0$ down to the recommended leaf $\ell_b$: $s_k \in (s_0, s_1,... , \ell_b)$, thus we have $\forall k \in \llbracket 0,D \rrbracket $
    $$\text{L}_{s_{k−1}}(t) = \max_{s\in \mathcal{C}(s_{k−1})} \text{L}_s(t) = \texttt{BAIReco}(s_{k−1}) = \text{L}_{s_{k}}$$
    which directly gives $\text{L}_{s_b}(\tau) = \text{L}_{\ell_b}(\tau)$.\\
    
    % \begin{proposition}
    % Assume that $\texttt{BLIstop} = \texttt{True}$, we have
        % $\forall s \in \mathcal{D}, \texttt{BAIStop}(s) = \texttt{True} \Leftrightarrow \forall s_a \in \text{Anc}(s), \texttt{BAIStop}(s_a) = \texttt{True}$
    % \end{proposition}
    
    The BAI problem is also solved for each ancestor of $\ell_b$, $\forall s \in \text{Anc}(\ell_b), \texttt{BAIStop}(s) = \texttt{True}$ and in particular for the node $s_p$, leading to $\text{U}_{s_c}(\tau) - \text{L}_{s_b}(\tau) \leq \varepsilon$ since \texttt{BAIStop}($s_p$) holds. To summarize we have
    \begin{equation}
    \nonumber
    \left\{
        \begin{split}
        &\text{U}_{s_c}(\tau) \geq \text{U}_{\ell_c}(\tau)\\ 
        &\text{L}_{s_b}(\tau) = \text{L}_{\ell_b}(\tau)
        \end{split}
    \right. \ \ \text{which imply} \ \ \text{U}_{\ell_c}(\tau) - \text{L}_{\ell_b}(\tau) \leq \varepsilon.
    \end{equation}
    
    Finally, from the Lemma.2 of \cite{kaufmann2017monte}, we obtain $\mathbb{P} \left[ V(\ell^*) - V(\ell_b) \leq \varepsilon \right] \geq 1 - \delta$ when the confidence intervals are calculated with the exploration function $\beta_1$ that we defined in section \ref{exploration_beta}.
    $$\beta_1(N,\delta) = \ln \left(\frac{|\mathcal{L}_\tau|}{\delta}\right) + 3 \ln \ln\left(\frac{|\mathcal{L}_\tau|}{\delta}\right) + \frac{3}{2} \ln \left(\ln(N) + 1 \right)$$
    
    \begin{flushright} $\blacksquare$ \end{flushright}

\end{document}